\documentclass[10pt,twocolumn,letterpaper]{article}
\usepackage{amsmath}
\DeclareMathOperator*{\argmin}{arg\,min} 

\usepackage[pagenumbers]{iccv} 

\usepackage{amsmath}

\definecolor{iccvblue}{rgb}{0.21,0.49,0.74}
\usepackage[pagebackref,breaklinks,colorlinks,allcolors=iccvblue]{hyperref}
\usepackage{multirow}
\usepackage{amsmath}
\def\paperID{8203} 
\def\confName{ICCV}
\def\confYear{2025}

\title{Multi-Label Stereo Matching for Transparent Scene Depth Estimation}

\author{Zhidan Liu\textsuperscript{\rm 1,2}\footnotemark[2],
Chengtang Yao\textsuperscript{\rm 1,2}\footnotemark[2],
Jiaxi Zeng\textsuperscript{\rm 1,2},
Yuwei Wu\textsuperscript{\rm 2,1},
Yunde Jia\textsuperscript{\rm 2,1},
 \\
\textsuperscript{\rm 1}Beijing Key Laboratory of Intelligent Information Technology, \\ School of Computer Science \& Technology, Beijing Institute of Technology, China \\
\textsuperscript{\rm 2}Guangdong Laboratory of Machine Perception and Intelligent Computing, \\
Shenzhen MSU-BIT University, China \\
{\tt\small \{zdliu,yao.c.t,jiaxi,wuyuwei,jiayunde\}@bit.edu.cn}
}

\begin{document}
\maketitle
\renewcommand{\thefootnote}{\fnsymbol{footnote}}
\footnotetext[2]{These authors contributed equally to this work.}
\begin{abstract}
In this paper, we present a multi-label stereo matching method to simultaneously estimate the depth of the transparent objects and the occluded background in transparent scenes.
Unlike previous methods that assume a unimodal distribution along the disparity dimension and formulate the matching as a single-label regression problem, we propose a multi-label regression formulation to estimate multiple depth values at the same pixel in transparent scenes. To resolve the multi-label regression problem, we introduce a pixel-wise multivariate Gaussian representation, where the mean vector encodes multiple depth values at the same pixel, and the covariance matrix determines whether a multi-label representation is necessary for a given pixel. The representation is iteratively predicted within a GRU framework. In each iteration, we first predict the update step for the mean parameters and then use both the update step and the updated mean parameters to estimate the covariance matrix. We also synthesize a dataset containing 10 scenes and 89 objects to validate the performance of transparent scene depth estimation. The experiments show that our method greatly improves the performance on transparent surfaces while preserving the background information for scene reconstruction. Code is available at \hyperlink{https://github.com/BFZD233/TranScene}{https://github.com/BFZD233/TranScene}.

\end{abstract}
    
\section{Introduction}

\begin{figure}[!h]
\centering
\includegraphics[width=0.48\textwidth]{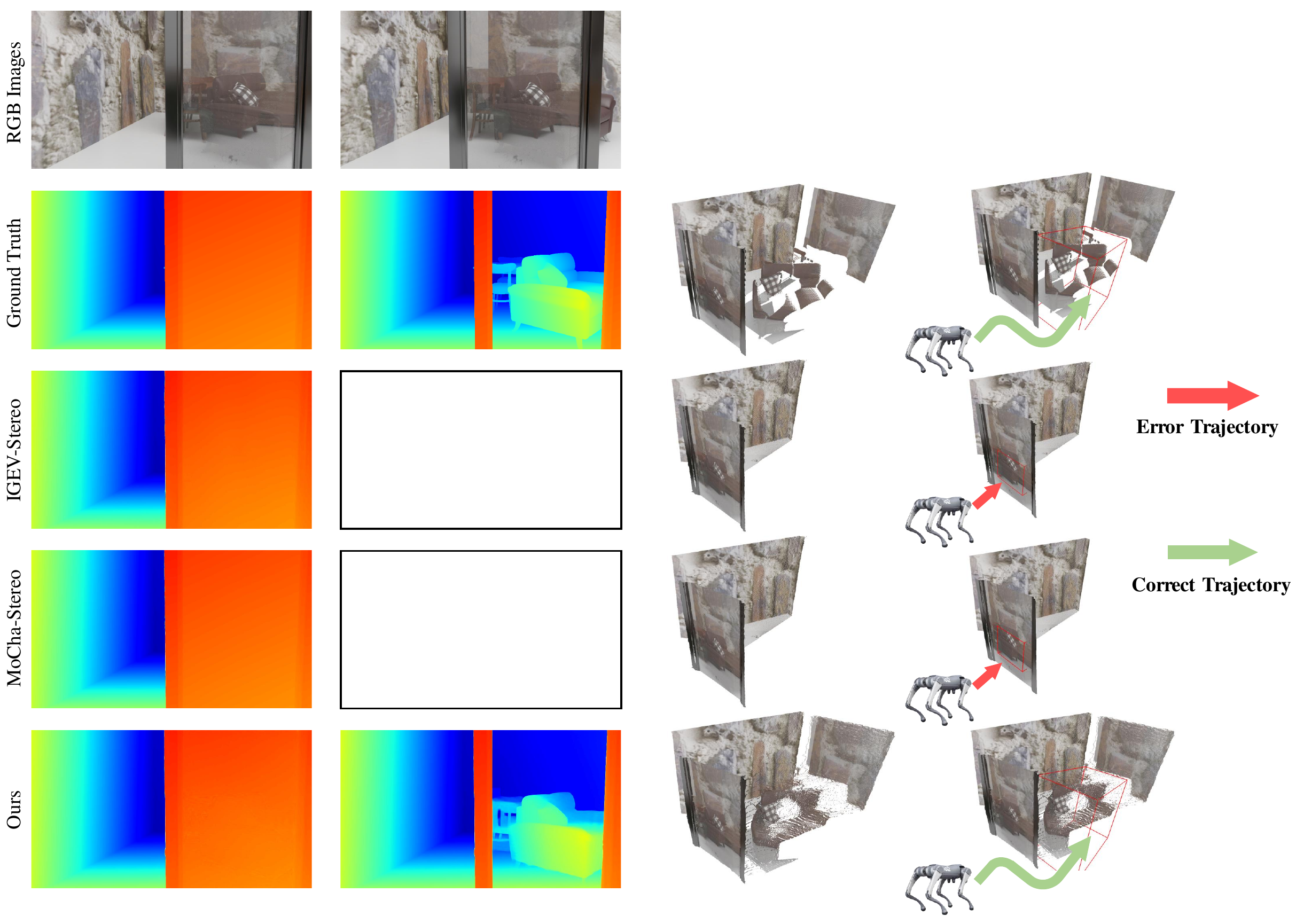}
\caption{The visualization of the multi-label disparity map and the reconstructed point cloud. The first column is the foreground disparity map and the second column is the background disparity map. The rest two columns represent the points cloud reconstructed by the multi-label disparity map and the corresponding robot navigation. Our method can estimate the depth of the transparent foreground and the occluded background simultaneously, as shown in the red and green box.}
\label{Fig: demo}\
\vspace{-0.7cm}
\end{figure}

Stereo matching aims to estimate a dense 3D scene geometry by matching points between a rectified image pair. It has various applications in scene understanding, Embodied AI, and AR/MR. In these downstream applications, a human wearing an AR/MR device or a robot is likely to interact with transparent scenes (e.g., glass windows, doors, and bottles). These transparent scenes require simultaneously estimating the depth of both the transparent objects and the background that is occluded by the transparent objects. Otherwise, a robot might get stuck in front of a glass door or fail to grasp objects behind a glass window, as shown in Figure \ref{Fig: demo}. However, existing methods mainly formulate the matching as a single-label regression problem and can only perceive one depth value for either the transparent objects or the occluded background, which is not suitable for transparent scenes, as illustrated in Figure \ref{Fig: demo}.

In this paper, we propose a multi-label stereo matching method for transparent scene depth estimation, e.g., glass windows and doors. In order to estimate the 3D geometry of both the transparent foreground and the occluded background, we formulate the matching process as a multi-label regression problem. In the context of multi-label stereo matching, the traditional single-value disparity map is inadequate, requiring a new representation that should satisfy the following three properties: (1) multiple labels should be allowed for a single pixel, (2) an indicator is needed to determine whether a pixel has only one or a limited number of labels, as multiple labels are only necessary for pixels on transparent surfaces, (3) the probability of each label corresponding to the transparent foreground and occluded background should individually approximate to 1, instead of summing to 1, as these labels represent distinct 3D objects respectively.

To this end, we propose a pixel-wise multivariate Gaussian representation parameterized by a mean vector and a covariance matrix. The mean vector allows multiple labels for a single pixel. The strictly lower/upper triangular parts of the covariance matrix capture the correlation between each pair of labels, indicating whether a pixel has only one or a limited number of labels. The marginal distribution of the multivariate Gaussian allows the probability of each label to individually approximate 1. In order to learn the multivariate Gaussian representation, we iteratively predict its parameters within a GRU framework. In each iteration, we first use independent convolution blocks for each label to predict the update step of the mean vector. We then use the updated mean vector to build a cost volume. This cost volume, along with the update step and left image features, is used to predict the covariance matrix. Notably, we parameterize the covariance matrix using the standard deviation and Pearson correlation coefficient, which reduces the number of predicted parameters.


\begin{figure*}
\centering
\includegraphics[width=0.95\textwidth]{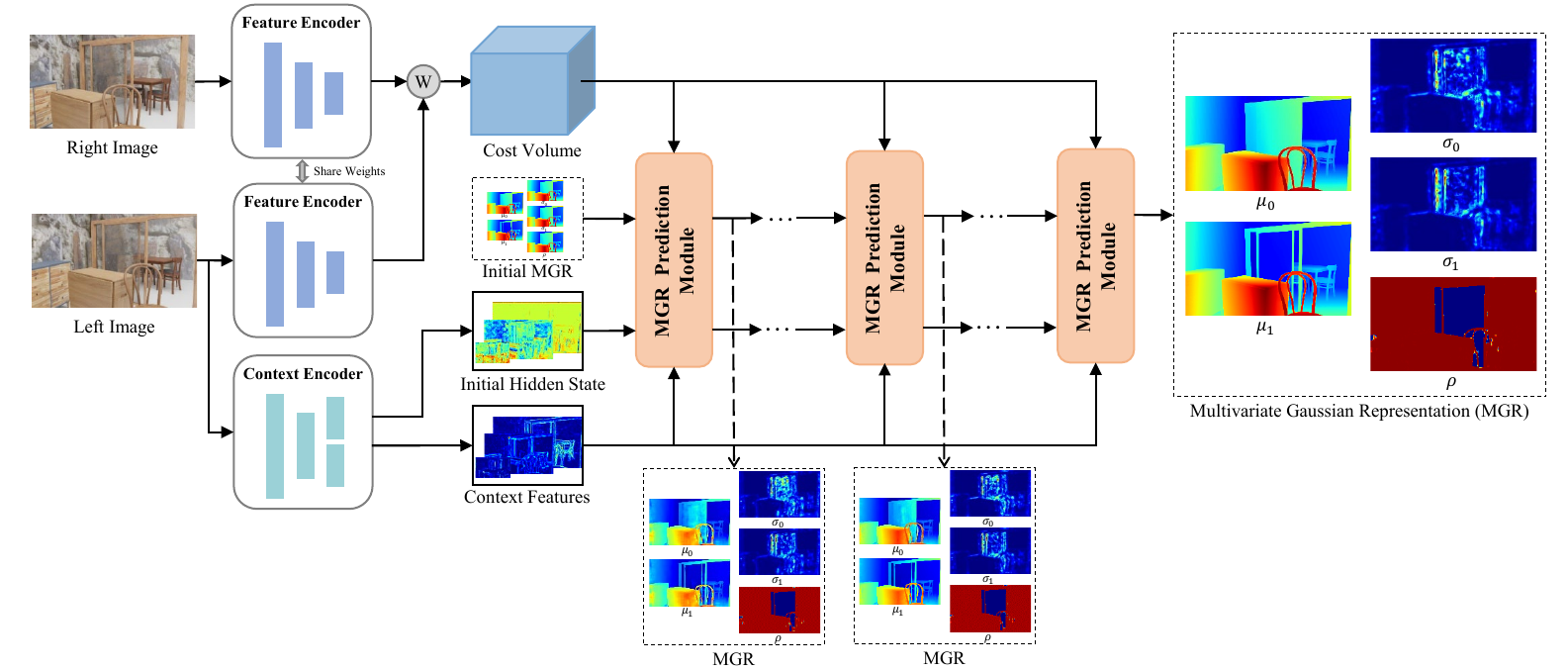}
\caption{The pipeline of our multi-label stereo matching method. Our multivariate Gaussian representation (MGR) is parameterized by a mean vector $\mu=(\mu_0,\mu_1)^T$ and a covariance matrix $\left[ \begin{smallmatrix} \sigma_0 & \rho\sigma_0\sigma_1 \\ \rho\sigma_0\sigma_1 & \sigma_1 \end{smallmatrix} \right]$, which are iteratively predicted in a GRU framework. \textcircled{w} represents the warping operation between left and right images. }
\label{Fig: pipeline}
\vspace{-0.4cm}
\end{figure*}

Existing public datasets are mostly captured in nontransparent scenes \cite{scharstein2014high,mayer2016large,schops2017multi,menze2015object}, while only a limited number of datasets consider transparent scene \cite{ramirez2022open}. These transparent scene datasets mainly focus on small objects (e.g., glass bottles), ignoring large furniture (e.g., glass doors and windows) which are very likely to appear when the robot/human interacts with the environments. In this paper, we create a photorealistic synthetic indoor dataset containing objects of various sizes and shapes to provide a good supplement to the current research mainstream. Our dataset is composed of 10k images with ground truth of multi-label disparity maps. We also provide semantic masks to distinguish between the transparent and the occluded areas. We collect 40 kinds of materials, 56 kinds of textures, and 89 kinds of objects to simulate various real-world scenes.

In the experiments, we compare our network with state-of-the-art methods in both synthetic and real-world datasets, transparent and occluded surfaces. The results show that our method greatly improves the reconstruction of transparent surfaces without sacrificing the performance on occluded objects. What's more, our method accurately reconstructs the foreground and background of transparent areas simultaneously.

Our main contributions are as follows:
\begin{itemize}
\item We propose a multi-label stereo matching method for transparent scene depth estimation by formulating the matching process as a multi-label regression problem.

\item We propose a pixel-wise multivariate Gaussian representation for multi-label stereo matching. The representation is parameterized by a mean vector and a covariance matrix, which is learned iteratively within a GRU framework.

\item We create a diverse synthetic dataset containing 10 kinds of scenes and 89 kinds of objects.

\end{itemize}

\section{Related Works}
\subsection{Deep Stereo Matching}
Deep stereo matching methods have achieved significant success by leveraging deep neural networks to learn the matching process. These methods can be broadly categorized into aggregation-based and iterative methods.
The aggregation-based methods first build a cost volume and then aggregate it using either CNN \cite{kendall2017end, chang2018pyramid, guo2019group, gu2020cascade, shen2021cfnet, yao2022foggystereo} or Transformer \cite{li2021revisiting, weinzaepfel2023croco}. 
The aggregated cost volume is subsequently converted into a probability volume through Softmax. The disparity map is later directly computed from the expectation of the probability distribution by assuming a unimodal distribution along the disparity dimension. While aggregation-based methods have made great progress, they struggle in ill-posed areas, such as textureless or low-texture areas. 
In contrast, iterative methods perform better in ill-posed areas by iteratively predicting the update step for disparity within a GRU framework \cite{lipson2021raft,li2022practical,xu2023iterative,zeng2023parameterized,liang2024any}. During each iteration, a multi-level GRU encodes the matching cost information and image semantic information into hidden states, from which the disparity update step is decoded. The iterative updation allows the predicted disparity to progressively converge towards the ground truth. Although these methods do not explicitly model a distribution along the disparity dimension, they implicitly approximate a single Gaussian distribution with a fixed variance \cite{zeng2023parameterized}.

The above methods have made significant progress, but their performance degrades noticeably when encountering scenes with transparent objects. These methods typically rely on a single-label disparity map and assume a unimodal distribution along the disparity dimension. This assumption mixes the network information flow, including matching information and semantic information. The mixed information flow is easy to make a negative impact on each other, while the disparity estimation on transparent surfaces depends on semantic information, and the estimation on non-transparent surfaces depends on matching information. Instead, our method disentangles the transparent and non-transparent surfaces through multi-label regression, improving the network information flow for different surfaces. Moreover, our multi-label stereo matching method enables the simultaneous reconstruction of both the transparent areas' foreground and background, enhancing scene interaction and understanding, particularly when a robot watches indoor/outdoor scenes through a glass window.

\subsection{Stereo Matching of Transparent Surface}
Stereo matching becomes challenging when dealing with transparent surfaces as the co-visibility of foreground and background makes the matching ambiguous. DDF \cite{chai2020deep} and CDP \cite{cai2023consistent} resolve the challenging by fusing the result of stereo matching with the output of the RGB-D camera. They achieve impressive performance but require additional hardware. DMStereo \cite{zhi2018deep} and ASGrasp \cite{shi2024asgrasp} use infrared (IR)  images to improve the reconstruction of transparent objects. However, IR imaging is only reliable in limited environments, as its image quality is often compromised by environmental lighting conditions. TA-Stereo \cite{wu2023transparent} employs an additional semantic mask to extract the transparent area and convert it into a non-transparent one by inpainting it with the texture of surrounding objects. TA-Stereo is effective but the sequential pipeline of semantic segmentation, image processing, and stereo matching is costly. Instead, some methods use pseudo labels to extend the dataset to unlabeled data and distill the knowledge from a teacher model to a student model \citet{costanzino2023learning,ramirez2024ntire,zhang2024robust}. They greatly improve the performance from the perspective of data, but rely on previous networks and thus still suffer from the limitations mentioned above, like ignoring the geometry of either foreground or background. Different from them, we propose a new framework to formulate the stereo matching in transparent scenes as a multi-label regression problem and introduce a multi-variate Gaussian representation. Our method resolves the challenge from a new view, which can be enhanced/integrated with the aforementioned methods, making it a valuable supplement to current research mainstream.

\begin{figure*}
\centering
\includegraphics[width=0.98\textwidth]{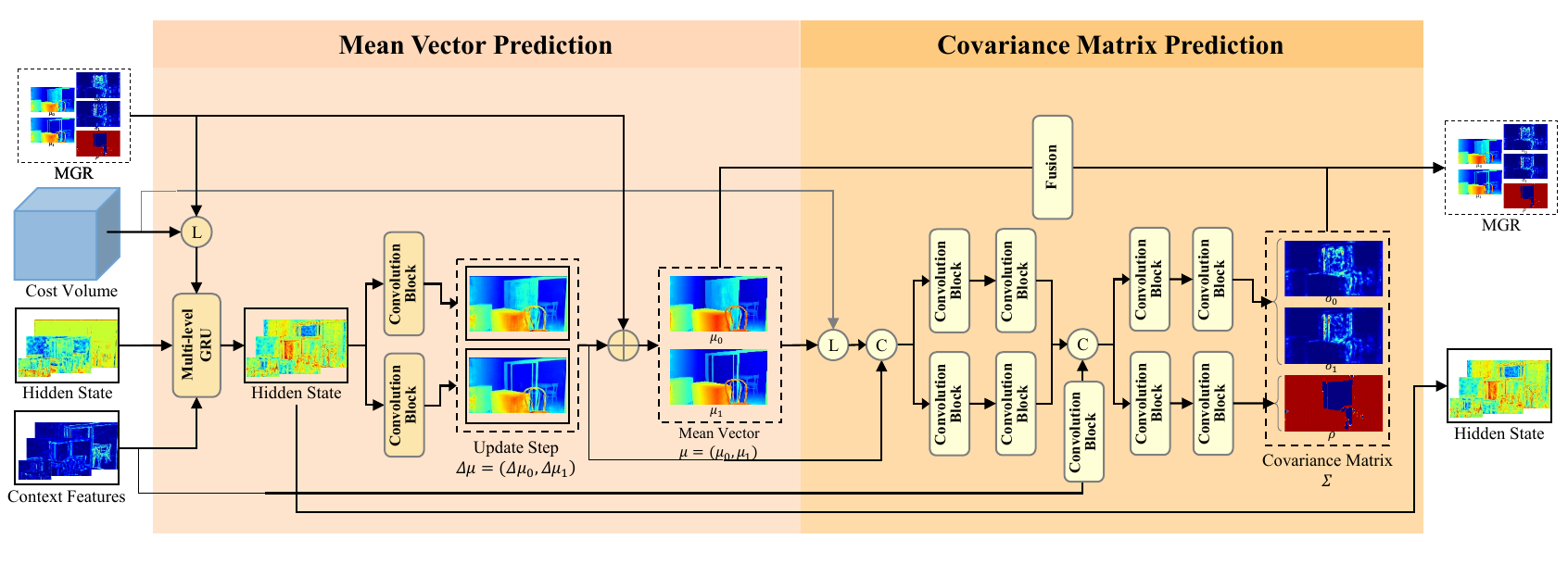}
\caption{The network architecture of our multivariate Gaussian representation (MGR) prediction module. we use a two-stage strategy to sequentially predict the mean vector $\mu=(\mu_0,\mu_1)^T$ and the covariance matrix $\left[ \begin{smallmatrix} \sigma_0^2 & \rho\sigma_0\sigma_1 \\ \rho\sigma_0\sigma_1 & \sigma_1^2 \end{smallmatrix} \right]$. \textcircled{\scalebox{0.7}{L}} is the lookup operation used to sample the cost value according to the predicted disparity. \textcircled{\scalebox{0.7}{C}} is the concatenation operation.}
\label{Fig: predcition}
\vspace{-0.4cm}
\end{figure*}
\section{Method}

\subsection{Problem Formulation}
Stereo matching is a pixel-wise labeling problem. Prior methods mainly formulate it as a single-label regression problem to achieve sub-pixel accuracy. They typically assume a unimodal distribution along the disparity dimension and compute the disparity $\hat{d}$ from the expectation of a probability distribution. Given a pair of images $\{ I_l, I_r \}$, the corresponding optimization objective is defined as follows:
\begin{equation}
\begin{aligned}
    &\theta = \argmin_{\theta} || \hat{d} - d ||, \\
    &where \  \hat{d} = \mathbb{E} \left( P(\hat{d} | I_l, I_r, C; \theta) \right).
\end{aligned}
\label{Eq: single-label}
\end{equation}
$d$ is the ground truth disparity map, $C$ is the cost volume constructed from $\{ I_l, I_r \}$, $\theta$ is the weights of the stereo matching networks, $\mathbb{E}(\cdot)$ is the expectation, and $P(\hat{d} | I_l, I_r; \theta)$ is the predicted or implicitly approximated distribution along the disparity dimension. The disparity estimation of transparent surfaces relies on semantic information, while that of non-transparent surfaces depends on matching information. However, the above formulation mixes the information flow of both semantic and matching information, leading to conflicts between them. Moreover, $P(\hat{d} | I_l, I_r; \theta)$ is typically supposed to be a unimodal distribution, which can result in the loss of geometric information in the foreground or background of transparent areas.

In this paper, we present a multi-label stereo matching method to accurately reconstruct the 3D geometry of both the transparent foreground and the non-transparent background simultaneously. We formulate stereo matching as a multi-label regression problem:
\begin{equation}
\begin{aligned}
    &\theta = \argmin_{\theta} \sum_{i=0}^{N-1}|| \hat{d}_i - d_i ||, \\
    &where \  \hat{d}_i = \mathbb{E}( P_m(\hat{d}_i | I_l, I_r, C; \theta) ).
\end{aligned}
\label{Eq: multi-label}
\end{equation}
Here, $N$ is the number of labels for each pixel, $\hat{d}_i$ and $d_i$ denote the predicted and ground-truth disparities at the $i$-th label respectively, and $P_m(\cdot)$ is the marginal distribution for each label. Our formulation assumes a multivariate distribution along the disparity dimension. Each variate corresponds to a label, representing a 3D object in the scene. We also assume a unimodal distribution for the marginal distribution of each label, where the probability of each label approximates 1, instead of summing to 1 like mixture distribution.

\subsection{Multivariate Gaussian Representation}
We introduce a pixel-wise multivariate Gaussian representation (MGR) to resolve the multi-label regression problem. The MGR enables (1) multiple labels for a single pixel, (2) an indication of each pixel's label count, and (3) the probability of each label to approximate 1. In this work, we mainly focus on the transparent property and assume the maximum number of labels is 2, such as the foreground and background of transparent areas. For more complicated scenes, our method can be seamlessly extended by increasing the number of labels. The MGR is parameterized by a mean vector $\mu$ and a covariance matrix $\Sigma$:
\begin{equation}
\begin{gathered}
    p(x;\mu,\Sigma) = \frac{1}{(2\pi)^{n/2}{|\Sigma|}^{1/2}} e^{ -\frac{1}{2}(x-\mu)^{T} \Sigma^{-1} (x - \mu) }, \\
    where \  
        \mu = 
        \begin{bmatrix}
        \mu_0 \\
        \mu_1
        \end{bmatrix}
        , \ 
        \Sigma =
        \begin{bmatrix}
            \sigma_0^2 & \rho \sigma_0 \sigma_1\\
            \rho \sigma_0 \sigma_1 & \sigma_1^2
        \end{bmatrix}.
\end{gathered}
\end{equation}
We use the Pearson correlation coefficient $\rho$ in $\Sigma$, where $\rho_{X,Y} = cov(X,Y) / (\sigma_X \sigma_Y)$, $-1 \leq \rho \leq 1$, $cov$ is covariance between two variables, $\sigma$ is the standard deviation. 
Here, $\rho$ can be further constrained by $0 \le \rho \le 1$ in multi-label stereo matching. 
In transparent scenes, $\rho$ is supposed to approximate 0 as the two labels are non-correlated, representing the foreground and background 
of transparent areas respectively. In this situation, $\mu_0$ and $\mu_1$ correspond to the disparity of foreground and background individually. In contrast, $\rho$ is supposed to approximate 1 in other normal areas, and there is only one label for each pixel. Thus, the $\mu_0$ and $\mu_1$ should be fused as the final result. This process can be described as
\begin{equation}
    \hat{d} = 
    \begin{cases}
        \mathbf{F}_f \left(\mu_0,\mu_1\right), \  & \rho \geq \alpha\\
        \{ \mu_0,\mu_1 \}, \  & \rho < \alpha 
    \end{cases},
\end{equation}
where $\mathbf{F}_f$ denotes the fusion process, $\alpha$ is a manual threshold to distinguish between transparent and normal areas.

\subsection{Network Architecture}
The pipeline of our network is illustrated in \ref{Fig: pipeline}. We extract image features from the left and right images to construct the cost volume, while context features are simultaneously extracted from the left image. The cost volume and context features are used to learn the parameters of multivariate Gaussian representation (MGR). The parameters are learned iteratively in our MGR prediction module, where a two-stage strategy is used to sequentially predict the mean vector and the covariance matrix in each iteration, as shown in Figure \ref{Fig: predcition}. In the first stage, we use a multi-level GRU to obtain the hidden state and decode the update step $\{ \Delta {\mu_0}, \Delta {\mu_1} \}$ for the mean vector $\{ \mu_0, \mu_1 \}$ from the hidden state. In the second stage, we use the updated mean vector and the context features to predict the covariance matrix $\{ \sigma_0, \sigma_1, \rho \}$. Finally, we fuse $\{ \mu_0, \mu_1 \}$ under the guidance of $\{ \sigma_0, \sigma_1, \rho \}$ for the disparity estimation in normal areas.

\subsubsection{Feature Extraction} 
Following RAFT-Stereo \cite{lipson2021raft}, we adopt the same architecture of feature encoder and context encoder. They are both multi-scale networks, including 1/4, 1/8, and 1/16 resolutions of the input images. The feature encoder extracts the image features of the left and right images through a series of residual blocks. The left and right image features are then used to construct the cost volume via the inner product. The context encoder has a similar architecture to the feature encoder but extracts context features solely from the left image. 

\subsubsection{Mean Vector Prediction} 
In the first stage of the MGR prediction module, we predict the update step $\Delta \mu = (\Delta \mu_0, \Delta\mu_1 )^T$ for each component of the mean vector $\mu = (\mu_0, \mu_1 )^T$. We first use a multi-level GRU to obtain a multi-resolution hidden state, including 1/4, 1/8, and 1/16 resolution of the input images.
The architecture of multi-level GRU follows prior methods \cite{lipson2021raft,xu2023iterative}. It takes the cost volume, the context features, and the hidden state from the last iteration as input, and outputs the hidden state at each resolution in turn by aggregating information from low resolution to high resolution. 
Next, we use a two-stream network to decode the update step for the mean vector from the hidden state at the 1/4 resolution. As shown in Figure \ref{Fig: pipeline}, each stream is independently built for each label via a convolution block. The block consists of four convolutions, each followed by a ReLU activation function except for the last one. Once we obtain the update step $\Delta \mu$, we add it to the mean vector from the last iteration.

\subsubsection{Covariance Matrix Prediction} 
In the second stage of the MGR prediction module, we predict the parameters of covariance matrix $\{\sigma_0, \sigma_1, \rho\}$ from the update step for the mean vector. We first use the updated mean vector to sample a thin cost volume for each label via look-up operation. The sampled cost volume, along with the update step and the context features, is then used to predict the covariance matrix. Since the channel number of the sampled cost volume and update step is much smaller than that of the context features, we concatenate them and use two independent streams for each label to increase the channel dimension of their concatenation. The outputs of these two streams are then concatenated with the context features that have been processed by two additional convolutions. This concatenation is used to predict $\{ \sigma_0, \sigma_1, \rho \}$ through two convolution blocks. The standard deviations $\{ \sigma_0, \sigma_1 \}$ are further activated by a softplus function, while the correlation coefficient $\rho$ is further activated by a sigmoid function.

\subsubsection{Fusion} 
There is only a single label for each pixel in normal areas, while multiple labels are redundant. Thus, we fuse $\mu_0$ and $\mu_1$ via weighted average to obtain the final result $\hat{d}$ for normal areas:
\begin{equation}
    \begin{gathered}
    \hat{d} = w^T \mu.
    \end{gathered}
\end{equation}
The weights $w = (w_0, w_1)^T$ are learned from the covariance matrix and the first concatenation in the covariance matrix prediction. 

\subsection{Loss Function}
We compute $l_1$ distance between the predicted mean vector $\mu$ and the multi-label ground truth $\mu_{gt}$:
\begin{equation}
    \mathcal{L}_{\mu} = \left\| \mu - \mu_{gt} \right\|_1.
\end{equation}
$\mu_{gt} = (\mu_{gt}^f, \mu_{gt}^b)^T$ for pixels in transparent areas,  $\mu_{gt} = (\mu_{gt}^f, \mu_{gt}^f)^T$ for pixels in normal areas, $\mu_{gt}^f$ is the disparity of foreground, and $\mu_{gt}^b$ is the disparity of background.

We also supervise the fusion result $\hat{d}$ with ground-truth single label disparity for normal areas:
    \begin{equation}
    \mathcal{L}_{d} = \left\| \hat{d} - u_{gt}^f \right\|_1.
    \end{equation}

To maximize the probability of the multi-label ground truth, we also minimize the negative log-likelihood loss:
\begin{equation}
\begin{aligned}
        \mathcal{L}_{lh} &= - \mathbb{E}_{\mu,\Sigma}\operatorname{log}p(x;\mu,\Sigma).
\end{aligned}
\end{equation}
Notably, we observe that the model convergence become unstable when optimizing the correlation coefficient $\rho$ through $\mathcal{L}_{lh}$. Thus, we independently supervise $\rho$ through
\begin{equation}
    \mathcal{L}_{\rho} = \left\| \rho - \rho_{gt} \right\|_1,
\end{equation}
where $\rho_{gt}$ is set as 0 for pixels on transparent surfaces and 0.95 for pixels on normal areas.

The final loss is the weighted sum of the above loss in each iteration:
\begin{equation}
\begin{gathered}
    \mathcal{L} = \Sigma_{i=1}^{M} \gamma^{M-i} (\mathcal{L}_{\mu}^i + \mathcal{L}_{d}^i + \beta_0\mathcal{L}_{lh}^i + \beta_1\mathcal{L}_{\rho}^i ),
\end{gathered}
\end{equation}
where $\beta_0$ and $\beta_1$ are hyperparameters used in tuning, $M$ is the number of iteration, $\gamma = 0.9$.

\subsection{Transparent Scene Dataset}
Our transparent scene dataset (TranScene) is a photorealistic synthetic indoor dataset composed of 10k images with multi-label ground truth. The multi-label ground truth is stored in the form of multiple disparity maps.  Our dataset also provides semantic masks to distinguish between transparent and non-transparent areas. 

\subsubsection{Dataset Creation} 
Following the Scene Flow dataset \cite{mayer2016large}, we use Blender to build up our dataset, including 10 types of scenes, 33 types of transparent objects, and 56 types of non-transparent objects. Unlike 
the Booster dataset \cite{ramirez2022open} limited to tabletops, our scenes range from room corners to tabletops and windows, simulating environments with which a human wearing an AR/MR device or a robot might interact. The layout of the scene is also carefully designed to simulate physically realistic settings.
The background of each scene is the wall with random textures. We randomly select $5 \sim 8$ non-transparent objects with specific materials and textures, along with 1 transparent object made of glass, and place them on the floor or tabletop. Additionally, we randomly apply a stain image as the roughness of the glass material to simulate dirt spots on the glass, as seen in real-world scenarios. In total, we have 40 materials and 56 textures. Each object is randomly rotated around the normal vector of its supporting surface to ensure contact with the floor or tabletop.
We set the camera baseline at 200 mm for room corners and 100 mm for tabletops, with a focal length of 933.34 mm for each scene. The camera is moved along 2 predefined 3D trajectories to capture frames of the scene. We render the disparity of each frame before and after removing the transparent objects. In total, we generate about 15000 stereo frames at a resolution of $540 \times 960$. 

\subsubsection{Evaluation Metrics}
Following the Booster dataset \cite{ramirez2022open}, we choose the endpoint error (EPE) and the percentage of pixels with an error larger than a threshold $\tau$ (bad-$\tau$) as disparity evaluation metrics. We choose the mean absolute error (MAE) and the percentage of pixels with an error distance larger than a threshold $\phi$ cm (bad-$\phi$) as depth evaluation metrics. We evaluate the performance of stereo models in the foreground of all areas (Foreground), the background of transparent areas (Background), and the entire areas (All).

\begin{figure*}
\centering
\includegraphics[width=0.9\textwidth]{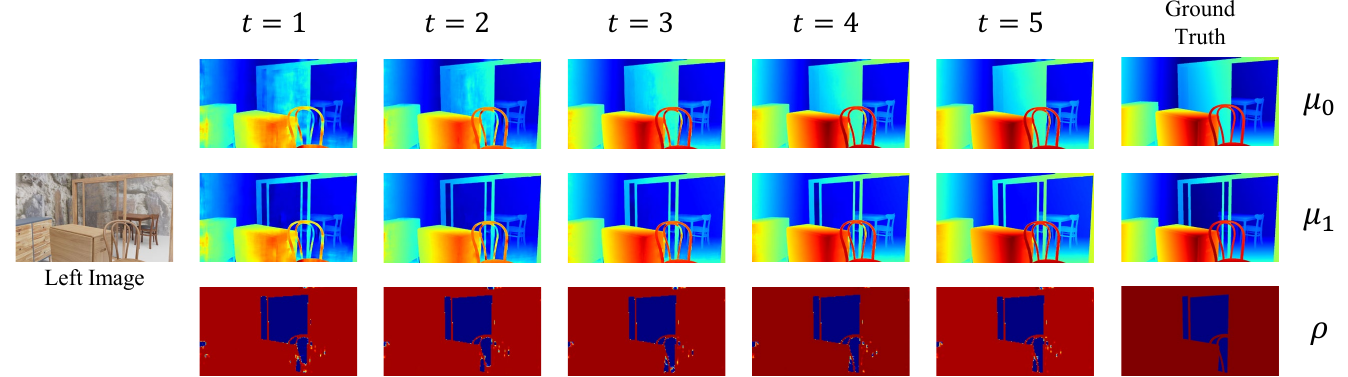}
\caption{The visualization of the parameters ($\mu_0,\mu_1,\rho$) of the multivariate Gaussian representation at each iteration.}
\label{Fig: iteration_parameters}
\vspace{-0.4cm}
\end{figure*}


\begin{table*}[h!]
\renewcommand\arraystretch{1.5}
    \centering
        \scalebox{0.8} {
    \begin{tabular}{c|cccc|cccc}
      \hline
      \multirow{2}{*}{Method} & \multicolumn{4}{c|}{ Foreground } & \multicolumn{4}{c}{ Background }  \\ \cline{2-9}
        &EPE $\downarrow$  &bad-2 $\downarrow$  &bad-3 $\downarrow$ &bad-5 $\downarrow$
        &EPE $\downarrow$  &bad-2 $\downarrow$  &bad-3 $\downarrow$ &bad-5 $\downarrow$ \\ \hline \hline
    $Baseline$ & 0.94  & 4.70 & 3.71 & 2.48  & - & - & - & - \\ \hline
    $ML+Baseline$ & 1.21 & 5.92 & 4.31 & 2.91 & 1.49 & 14.68 & 10.28 & 6.34\\ \hline
    $MGR+Baseline$ & 0.94 & 5.17 & 3.62 & 2.44 & 1.56 & 14.42 & 10.16 & 6.55\\  \hline
    $MGR+\mu$ & 0.96 & 4.62 & 3.37 & 2.29 & 1.39 & 12.62 & 9.10 & 5.39\\ \hline
    $MGR+\mu+SAM$ & 0.81 & 3.76 & 2.82 & 1.92 & 1.01 & 9.09 & 6.56 & 4.40\\ \hline
    \end{tabular}
    }
    \caption{The ablation study of each configuration for our model on the TranScene datatset. $Baseline$ is RAFT-Stereo. $MGR+Baseline$ is the basic architecture of our model. $MGR+\mu$ is $MGR+Baseline$ with two-stream network for mean vector prediction. $MGR+\mu+Conv$ is $MGR+\mu$ with additional 2 convolutional layers in each steam.  $MGR+\mu+SAM$ directly estimates multi-label disparity without leveraging MGR.. $MGR+\mu+SAM$ is use SAM to replace the original feature extractor. }
    \label{tab: components_study}
\vspace{-0.4cm}
\end{table*}

\begin{table}[!h]
\renewcommand\arraystretch{1.5}
    \centering
    \scalebox{0.8} {
    \begin{tabular}{c|cccc}
      \hline
      \multirow{2}{*}{Method} & \multicolumn{4}{c}{ Foreground }   \\ \cline{2-5}
        &EPE $\downarrow$  &bad-2 $\downarrow$  &bad-3 $\downarrow$ &bad-5 $\downarrow$
         \\ \hline \hline
    RAFT-Stereo\cite{lipson2021raft} & 0.94  & 4.70 & 3.71 & 2.48   \\ \hline
    IGEV-Stereo\cite{xu2023iterative} & 3.22 & 8.81 & 6.95 & 5.36\\ \hline
    Select-RAFT\cite{Wang_2024_CVPR} & 1.24 & 5.55 & 4.51 & 3.44  \\ \hline
    Select-IGEV\cite{Wang_2024_CVPR} & 1.29 & 3.39 & 2.73 & 1.95\\ \hline
    MoCha-Stereo\cite{Chen_2024_CVPR} & 2.83 & 5.55 & 4.92 & 4.30\\ \hline
    Ours & 0.96 & 4.62 & 3.37 & 2.29  \\ \hline
    Ours + SAM & 0.81 & 3.76 & 2.82 & 1.92 \\ \hline
    \end{tabular}
    }
    \caption{The disparity comparison of algorithms on the TranScene dataset.}
    \label{tab: comparision}
    \vspace{-0.6cm}

\end{table}

\section{Experiments}
\subsection{Datasets}
\subsubsection{Scene Flow} 
Scene Flow \cite{mayer2016large} is a synthetic dataset including 35454 training and 4370 testing stereo pairs with a resolution of $540 \times 960$. It provides dense ground truth disparity maps rendered from synthetic scenes. 

\subsubsection{Booster} 
Booster \cite{ramirez2022open} is a real-world dataset containing specular and transparent objects with an image resolution of $3008 \times 4112$. It mainly focuses on the tabletop scenario and provides ground truth disparity and material segmentation masks. There are 228 stereo pairs for training and 191 stereo pairs for testing.

\subsubsection{TranScene}
TranScene is a synthetic indoor dataset containing multi-label ground truth and transparent semantic masks. The dataset contains 10 kinds of scenes ranging from room corners to tabletops and windows. There are 11770 and 1422 image pairs with a resolution of $540 \times 960$ for training and testing respectively.

\begin{figure*}
\includegraphics[width=0.98\textwidth]{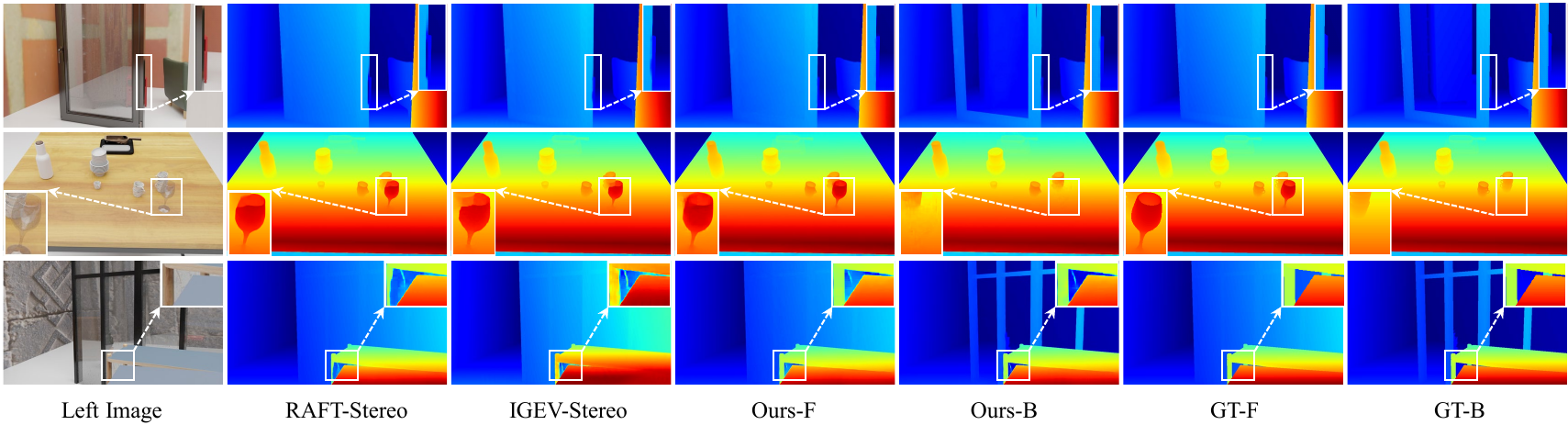}
\caption{The visualization of the disparity map on TranScene dataset.}
\label{Fig: vis TranScene}
\end{figure*}

\subsection{Implementation details}
Our model is based on the implementation of RAFT-Stereo \cite{lipson2021raft}. We set the iteration $M$ as 22 in training and 32 in testing. During testing, our model only conducts the covariance matrix prediction at the last iteration, achieving fast inference speed. During training, we use the AdamW as the optimizer and a one-cycle learning rate schedule with a maximum learning rate of 0.0002. We also apply the data augmentation used in RAFT-Stereo to the training of our method on all datasets. In the Scene Flow dataset, we train our model with a batch size of 8 for 200k steps on a single label. In the TranScene dataset, we train our model with a batch size of 8 for 200k steps first and then train the fusion model separately. As for the Booster dataset, we also use the model pre-trained in the TranScene dataset. For the hyperparameters, we set $\alpha=0.5, \gamma=0.9, \beta_0=0.00001, \beta_1=1$, respectively.

The predicted multi-variate Gaussian representation is at 1/4 resolution of the input image. We upsample it to full resolution using a convex combination of the $3 \times 3$ grid of their coarse resolution neighbors. The convex combination weights from the hidden state at 1/4 resolution via three regression networks for $\mu_0$, $\mu_1$, and $\Sigma$ respectively.

\subsection{Ablation Study}
We analyze the effectiveness of our multivariate Gaussian representation (MGR) in different configurations. As shown in table \ref{tab: components_study}, the model using MGR can predict the disparity of the transparent objects' foreground and background simultaneously while preserving comparable results on non-transparent objects. After using a two-stream network for mean vector prediction, instead of sharing the same convolutions for all labels in the mean vector, our model achieves a better performance in the background of transparent areas. 


When we add more convolutions into each stream, the performance deteriorates. We replaced the context feature extractor in RAFT-Stereo with the pretrained SAM2 model\cite{ravi2024sam2} to obtain stronger context information. The results demonstrate that incorporating stronger context features leads to a significant performance improvement. When we directly estimate multi-label disparity without our MGR, Table \ref{tab: components_study} shows a noticeable performance decline.


\begin{table}[!h]
\renewcommand\arraystretch{1.5}
    \centering
    \scalebox{0.7} {
    \begin{tabular}{c|ccccc}
      \hline
      \multirow{2}{*}{Method} & \multicolumn{5}{c}{ All }   \\ \cline{2-6}
        &MAE.(m) $\downarrow$  &bad-3 $\downarrow$  &bad-5 $\downarrow$ &bad-7 $\downarrow$ &bad-10 $\downarrow$
         \\ \hline \hline
    RAFT-Stereo\cite{lipson2021raft} & 0.188  & 27.18 & 25.75 & 25.12 & 24.51   \\ \hline
    IGEV-Stereo\cite{xu2023iterative} & 0.189 & 29.27 & 26.37 & 25.50 &24.81\\ \hline
    Select-RAFT\cite{Wang_2024_CVPR}  &0.188 &27.66 &26.13 &25.41 &24.85  \\ \hline
    Select-IGEV\cite{Wang_2024_CVPR}  &0.186 &26.08 &25.22 &24.82 &24.36\\ \hline
    MoCha-Stereo\cite{Chen_2024_CVPR}  &0.187 &26.40 &25.46 &25.01 &24.51\\ \hline
    Ours &\textbf{0.014} &\textbf{7.24} &\textbf{3.52} &\textbf{2.22} &\textbf{1.38} \\ \hline
    \end{tabular}
    }
    \caption{The depth comparison of algorithms on the TranScene dataset.}
    \label{tab: comparision_depth}
\vspace{-0.3cm}
\end{table}

We also provide the visualization of the parameters of the multivariate Gaussian representation at each iteration in Figure \ref{Fig: iteration_parameters}. As the number of iterations increases, the disparity of the transparent areas' foreground becomes smoother, while the disparity of background becomes clearer. The noise in $\rho$ is also gradually filtered out.


\begin{figure*}
\centering
\includegraphics[width=0.98\textwidth]{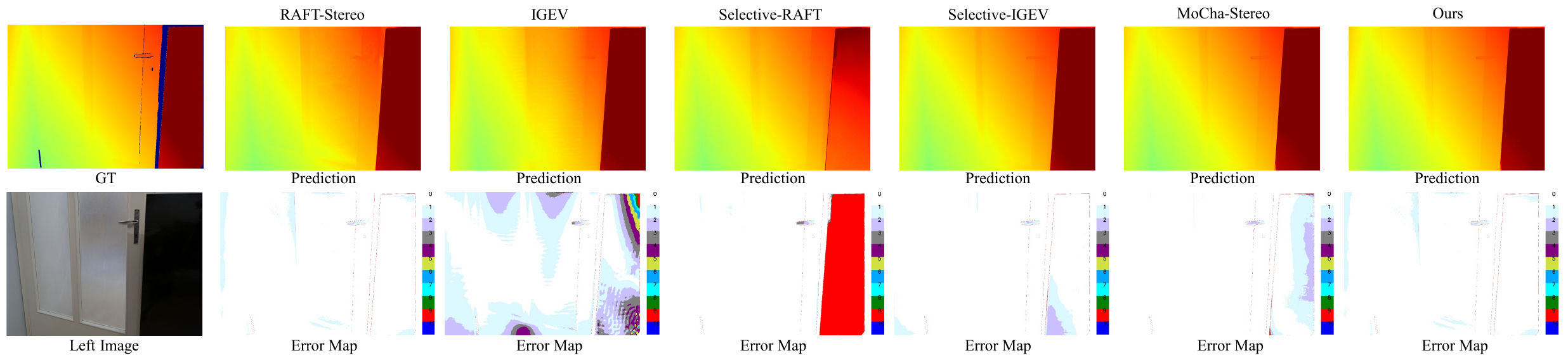}
\caption{The visualization of the disparity map and error map on Booster dataset.}
\label{Fig: booster}

\end{figure*}

\subsection{Comparisons with State-of-the-art Methods}

\textbf{TranScene} 
As shown in Table \ref{tab: comparision}, we achieve the best result in the foreground of all areas. Furthermore, as shown in Figure \ref{Fig: vis TranScene},  our method can reconstruct the background simultaneously and perform well on object edges. To assess the completeness of the entire scene geometric reconstruction, we evaluate in depth space. As shown in Table \ref{tab: comparision_depth}, our method achieves significant improvements.

We also find that RAFT-Stereo is smoother and much better in the planar region, but it will fail in thin objects. IGEV-Stereo performs well in small objects and provides better identification for transparent surfaces. However, it struggles with disparity estimation in texture-less regions and transparent surfaces.

\textbf{Booster} 
We evaluate the model trained on the TranScene dataset using the training set of the Booster dataset at 1/4 resolution. As shown in Table \ref{tab: comparision_booster}, all the models show a significant improvement compared to RAFT-Stereo trained on the Scene Flow dataset in the transparent region. Since our task focuses more on depth estimation in transparent scenes, while the Booster dataset contains a large number of object-level transparent objects, our method does not show significant improvements. As shown in the Figure \ref{Fig: booster}, our method performs well in transparent scenes.

\subsection{Limitations and Discussion}
As aforementioned, our method has achieved great performance in the reconstruction of both transparent areas' foreground and background. However, we also find some limitations of our model. In multi-label stereo matching, the data of transparent areas are limited, thus the long-tail problem or label balance problem becomes more and more important. Besides, our correlation $\rho$ works like a semantic map and still has some artifacts, a more powerful network may be considered to be injected into our architecture. We currently only consider two labels, but our model can be extended to more labels for complicated scenes, like reconstruction from the reflection in the mirror.

\begin{table}[h!]
\renewcommand\arraystretch{1.2}
    \centering
    \scalebox{0.8} {
    \begin{tabular}{c|cc|cc}
      \hline
      \multirow{2}{*}{Method} & \multicolumn{2}{c|}{ Trans } & \multicolumn{2}{c}{ NonTrans }  \\ \cline{2-5}
        &EPE $\downarrow$  &bad-3 $\downarrow$ &EPE $\downarrow$ &bad-3 $\downarrow$ \\ \hline \hline
    RAFT-Stereo*\cite{lipson2021raft} & 7.98  & 35.90 & 1.11  & 4.34  \\ \hline
    RAFT-Stereo\cite{lipson2021raft} & 5.10  & 26.35 & 1.26  & 7.37  \\ \hline
    IGEV-Stereo\cite{xu2023iterative} & 7.26 & 36.72 & 3.69 & 21.13 \\ \hline
    Select-RAFT\cite{Wang_2024_CVPR} &6.09 &27.01 & 1.98 & 8.35\\ \hline
    Select-IGEV\cite{Wang_2024_CVPR} &4.76 & 22.13 & 1.24 & 4.30\\ \hline
    MoCha-Stereo\cite{Chen_2024_CVPR}&3.87 & 20.43 & 1.14 & 3.47\\ \hline
    Ours & 5.33 &30.09 & 1.04 & 3.16\\ \hline
    \end{tabular}
    }
    \caption{The comparison of algorithms on the Booster dataset. RAFT-Stereo* is trained on the Scene Flow dataset.}
    \label{tab: comparision_booster}
\end{table}

\section{Conclusion}
In this paper, we have presented a multi-label stereo matching method that can estimate the 3D geometry of the transparent foreground and the occluded background simultaneously. We formulated the stereo matching as a multi-label regression problem and solved it by a multivariate Gaussian representation parameterized by a mean vector and a covariance matrix. The representation is efficiently learned in an iterative GRU framework, which enables the disentanglement of mixed matching information and semantic information. The experiments demonstrate that our method not only can achieve state-of-the-art performance in normal areas and the transparent foreground, but also preserve accurate 3D geometry of the occluded background.
{
    \small
    \bibliographystyle{ieeenat_fullname}
    \bibliography{main}

\begin{thebibliography}{29}
\providecommand{\natexlab}[1]{#1}
\providecommand{\url}[1]{\texttt{#1}}
\expandafter\ifx\csname urlstyle\endcsname\relax
  \providecommand{\doi}[1]{doi: #1}\else
  \providecommand{\doi}{doi: \begingroup \urlstyle{rm}\Url}\fi

\bibitem[Cai et~al.(2023)Cai, Zhu, Zhang, and Ren]{cai2023consistent}
Yuxiang Cai, Yifan Zhu, Haiwei Zhang, and Bo Ren.
\newblock Consistent depth prediction for transparent object reconstruction from rgb-d camera.
\newblock In \emph{Proceedings of the IEEE International Conference on Computer Vision}, pages 3459--3468, 2023.

\bibitem[Chai et~al.(2020)Chai, Wu, and Tsao]{chai2020deep}
Chun-Yu Chai, Yu-Po Wu, and Shiao-Li Tsao.
\newblock Deep depth fusion for black, transparent, reflective and texture-less objects.
\newblock In \emph{2020 IEEE International Conference on Robotics and Automation}, pages 6766--6772. IEEE, 2020.

\bibitem[Chang and Chen(2018)]{chang2018pyramid}
Jia-Ren Chang and Yong-Sheng Chen.
\newblock Pyramid stereo matching network.
\newblock In \emph{Proceedings of the IEEE Conference on Computer Vision and Pattern Recognition}, pages 5410--5418, 2018.

\bibitem[Chen et~al.(2024)Chen, Long, Yao, Zhang, Wang, Qin, and Wu]{Chen_2024_CVPR}
Ziyang Chen, Wei Long, He Yao, Yongjun Zhang, Bingshu Wang, Yongbin Qin, and Jia Wu.
\newblock Mocha-stereo: Motif channel attention network for stereo matching.
\newblock In \emph{Proceedings of the IEEE Conference on Computer Vision and Pattern Recognition}, pages 27768--27777, 2024.

\bibitem[Costanzino et~al.(2023)Costanzino, Ramirez, Poggi, Tosi, Mattoccia, and Di~Stefano]{costanzino2023learning}
Alex Costanzino, Pierluigi~Zama Ramirez, Matteo Poggi, Fabio Tosi, Stefano Mattoccia, and Luigi Di~Stefano.
\newblock Learning depth estimation for transparent and mirror surfaces.
\newblock In \emph{Proceedings of the IEEE International Conference on Computer Vision}, pages 9244--9255, 2023.

\bibitem[Gu et~al.(2020)Gu, Fan, Zhu, Dai, Tan, and Tan]{gu2020cascade}
Xiaodong Gu, Zhiwen Fan, Siyu Zhu, Zuozhuo Dai, Feitong Tan, and Ping Tan.
\newblock Cascade cost volume for high-resolution multi-view stereo and stereo matching.
\newblock In \emph{Proceedings of the IEEE Conference on Computer Vision and Pattern Recognition}, pages 2495--2504, 2020.

\bibitem[Guo et~al.(2019)Guo, Yang, Yang, Wang, and Li]{guo2019group}
Xiaoyang Guo, Kai Yang, Wukui Yang, Xiaogang Wang, and Hongsheng Li.
\newblock Group-wise correlation stereo network.
\newblock In \emph{Proceedings of the IEEE Conference on Computer Vision and Pattern Recognition}, pages 3273--3282, 2019.

\bibitem[Kendall et~al.(2017)Kendall, Martirosyan, Dasgupta, Henry, Kennedy, Bachrach, and Bry]{kendall2017end}
Alex Kendall, Hayk Martirosyan, Saumitro Dasgupta, Peter Henry, Ryan Kennedy, Abraham Bachrach, and Adam Bry.
\newblock End-to-end learning of geometry and context for deep stereo regression.
\newblock In \emph{Proceedings of the IEEE International Conference on Computer Vision}, pages 66--75, 2017.

\bibitem[Li et~al.(2022)Li, Wang, Xiong, Cai, Yan, Yang, Liu, Fan, and Liu]{li2022practical}
Jiankun Li, Peisen Wang, Pengfei Xiong, Tao Cai, Ziwei Yan, Lei Yang, Jiangyu Liu, Haoqiang Fan, and Shuaicheng Liu.
\newblock Practical stereo matching via cascaded recurrent network with adaptive correlation.
\newblock In \emph{Proceedings of the IEEE Conference on Computer Vision and Pattern Recognition}, pages 16263--16272, 2022.

\bibitem[Li et~al.(2021)Li, Liu, Drenkow, Ding, Creighton, Taylor, and Unberath]{li2021revisiting}
Zhaoshuo Li, Xingtong Liu, Nathan Drenkow, Andy Ding, Francis~X Creighton, Russell~H Taylor, and Mathias Unberath.
\newblock Revisiting stereo depth estimation from a sequence-to-sequence perspective with transformers.
\newblock In \emph{Proceedings of the IEEE International Conference on Computer Vision}, pages 6197--6206, 2021.

\bibitem[Liang and Li(2024)]{liang2024any}
Zhaohuai Liang and Changhe Li.
\newblock Any-stereo: Arbitrary scale disparity estimation for iterative stereo matching.
\newblock In \emph{Proceedings of the AAAI Conference on Artificial Intelligence}, pages 3333--3341, 2024.

\bibitem[Lipson et~al.(2021)Lipson, Teed, and Deng]{lipson2021raft}
Lahav Lipson, Zachary Teed, and Jia Deng.
\newblock Raft-stereo: Multilevel recurrent field transforms for stereo matching.
\newblock In \emph{2021 International Conference on 3D Vision}, pages 218--227. IEEE, 2021.

\bibitem[Mayer et~al.(2016)Mayer, Ilg, Hausser, Fischer, Cremers, Dosovitskiy, and Brox]{mayer2016large}
Nikolaus Mayer, Eddy Ilg, Philip Hausser, Philipp Fischer, Daniel Cremers, Alexey Dosovitskiy, and Thomas Brox.
\newblock A large dataset to train convolutional networks for disparity, optical flow, and scene flow estimation.
\newblock In \emph{Proceedings of the IEEE Conference on Computer Vision and Pattern Recognition}, pages 4040--4048, 2016.

\bibitem[Menze and Geiger(2015)]{menze2015object}
Moritz Menze and Andreas Geiger.
\newblock Object scene flow for autonomous vehicles.
\newblock In \emph{Proceedings of the IEEE Conference on Computer Vision and Pattern Recognition}, pages 3061--3070, 2015.

\bibitem[Ramirez et~al.(2022)Ramirez, Tosi, Poggi, Salti, Mattoccia, and Di~Stefano]{ramirez2022open}
Pierluigi~Zama Ramirez, Fabio Tosi, Matteo Poggi, Samuele Salti, Stefano Mattoccia, and Luigi Di~Stefano.
\newblock Open challenges in deep stereo: the booster dataset.
\newblock In \emph{Proceedings of the IEEE Conference on Computer Vision and Pattern Recognition}, pages 21168--21178, 2022.

\bibitem[Ramirez et~al.(2024)Ramirez, Tosi, Di~Stefano, Timofte, Costanzino, Poggi, Salti, Mattoccia, Zhang, Wu, et~al.]{ramirez2024ntire}
Pierluigi~Zama Ramirez, Fabio Tosi, Luigi Di~Stefano, Radu Timofte, Alex Costanzino, Matteo Poggi, Samuele Salti, Stefano Mattoccia, Yangyang Zhang, Cailin Wu, et~al.
\newblock Ntire 2024 challenge on hr depth from images of specular and transparent surfaces.
\newblock In \emph{Proceedings of the IEEE Conference on Computer Vision and Pattern Recognition}, pages 6499--6512, 2024.

\bibitem[Ravi et~al.(2024)Ravi, Gabeur, Hu, Hu, Ryali, Ma, Khedr, R{\"a}dle, Rolland, Gustafson, Mintun, Pan, Alwala, Carion, Wu, Girshick, Doll{\'a}r, and Feichtenhofer]{ravi2024sam2}
Nikhila Ravi, Valentin Gabeur, Yuan-Ting Hu, Ronghang Hu, Chaitanya Ryali, Tengyu Ma, Haitham Khedr, Roman R{\"a}dle, Chloe Rolland, Laura Gustafson, Eric Mintun, Junting Pan, Kalyan~Vasudev Alwala, Nicolas Carion, Chao-Yuan Wu, Ross Girshick, Piotr Doll{\'a}r, and Christoph Feichtenhofer.
\newblock Sam 2: Segment anything in images and videos.
\newblock \emph{arXiv preprint arXiv:2408.00714}, 2024.

\bibitem[Scharstein et~al.(2014)Scharstein, Hirschm{\"u}ller, Kitajima, Krathwohl, Ne{\v{s}}i{\'c}, Wang, and Westling]{scharstein2014high}
Daniel Scharstein, Heiko Hirschm{\"u}ller, York Kitajima, Greg Krathwohl, Nera Ne{\v{s}}i{\'c}, Xi Wang, and Porter Westling.
\newblock High-resolution stereo datasets with subpixel-accurate ground truth.
\newblock In \emph{German Conference on Pattern Recognition)}, pages 31--42, 2014.

\bibitem[Schops et~al.(2017)Schops, Schonberger, Galliani, Sattler, Schindler, Pollefeys, and Geiger]{schops2017multi}
Thomas Schops, Johannes~L Schonberger, Silvano Galliani, Torsten Sattler, Konrad Schindler, Marc Pollefeys, and Andreas Geiger.
\newblock A multi-view stereo benchmark with high-resolution images and multi-camera videos.
\newblock In \emph{Proceedings of the IEEE Conference on Computer Vision and Pattern Recognition}, pages 3260--3269, 2017.

\bibitem[Shen et~al.(2021)Shen, Dai, and Rao]{shen2021cfnet}
Zhelun Shen, Yuchao Dai, and Zhibo Rao.
\newblock Cfnet: Cascade and fused cost volume for robust stereo matching.
\newblock In \emph{Proceedings of the IEEE Conference on Computer Vision and Pattern Recognition}, pages 13906--13915, 2021.

\bibitem[Shi et~al.(2024)Shi, Jin, Li, Niu, Jin, Wang, et~al.]{shi2024asgrasp}
Jun Shi, Yixiang Jin, Dingzhe Li, Haoyu Niu, Zhezhu Jin, He Wang, et~al.
\newblock Asgrasp: Generalizable transparent object reconstruction and grasping from rgb-d active stereo camera.
\newblock In \emph{IEEE International Conference on Robotics and Automation}, 2024.

\bibitem[Wang et~al.(2024)Wang, Xu, Jia, and Yang]{Wang_2024_CVPR}
Xianqi Wang, Gangwei Xu, Hao Jia, and Xin Yang.
\newblock Selective-stereo: Adaptive frequency information selection for stereo matching.
\newblock In \emph{Proceedings of the IEEE Conference on Computer Vision and Pattern Recognition}, pages 19701--19710, 2024.

\bibitem[Weinzaepfel et~al.(2023)Weinzaepfel, Lucas, Leroy, Cabon, Arora, Br{\'e}gier, Csurka, Antsfeld, Chidlovskii, and Revaud]{weinzaepfel2023croco}
Philippe Weinzaepfel, Thomas Lucas, Vincent Leroy, Yohann Cabon, Vaibhav Arora, Romain Br{\'e}gier, Gabriela Csurka, Leonid Antsfeld, Boris Chidlovskii, and J{\'e}r{\^o}me Revaud.
\newblock Croco v2: Improved cross-view completion pre-training for stereo matching and optical flow.
\newblock In \emph{Proceedings of the IEEE/CVF International Conference on Computer Vision}, pages 17969--17980, 2023.

\bibitem[Wu et~al.(2023)Wu, Su, Chen, and Fan]{wu2023transparent}
Zhiyuan Wu, Shuai Su, Qijun Chen, and Rui Fan.
\newblock Transparent objects: A corner case in stereo matching.
\newblock In \emph{2023 IEEE International Conference on Robotics and Automation}, pages 12353--12359. IEEE, 2023.

\bibitem[Xu et~al.(2023)Xu, Wang, Ding, and Yang]{xu2023iterative}
Gangwei Xu, Xianqi Wang, Xiaohuan Ding, and Xin Yang.
\newblock Iterative geometry encoding volume for stereo matching.
\newblock In \emph{Proceedings of the IEEE Conference on Computer Vision and Pattern Recognition}, pages 21919--21928, 2023.

\bibitem[Yao and Yu(2022)]{yao2022foggystereo}
Chengtang Yao and Lidong Yu.
\newblock Foggystereo: Stereo matching with fog volume representation.
\newblock In \emph{Proceedings of the IEEE/CVF Conference on Computer Vision and Pattern Recognition}, pages 13043--13052, 2022.

\bibitem[Zeng et~al.(2023)Zeng, Yao, Yu, Wu, and Jia]{zeng2023parameterized}
Jiaxi Zeng, Chengtang Yao, Lidong Yu, Yuwei Wu, and Yunde Jia.
\newblock Parameterized cost volume for stereo matching.
\newblock In \emph{Proceedings of the IEEE International Conference on Computer Vision}, pages 18347--18357, 2023.

\bibitem[Zhang et~al.(2024)Zhang, Li, Huang, Yu, Gu, Zheng, and Bai]{zhang2024robust}
Jiawei Zhang, Jiahe Li, Lei Huang, Xiaohan Yu, Lin Gu, Jin Zheng, and Xiao Bai.
\newblock Robust synthetic-to-real transfer for stereo matching.
\newblock In \emph{Proceedings of the IEEE Conference on Computer Vision and Pattern Recognition}, pages 20247--20257, 2024.

\bibitem[Zhi et~al.(2018)Zhi, Pires, Hebert, and Narasimhan]{zhi2018deep}
Tiancheng Zhi, Bernardo~R Pires, Martial Hebert, and Srinivasa~G Narasimhan.
\newblock Deep material-aware cross-spectral stereo matching.
\newblock In \emph{Proceedings of the IEEE Conference on Computer Vision and Pattern Recognition}, pages 1916--1925, 2018.

\end{thebibliography}
}

\end{document}